\documentclass[10pt]{article}

\usepackage{subcaption}

\usepackage{geometry}                
\usepackage{xspace}
\usepackage{amssymb}
\usepackage{amsthm}
\usepackage{multirow}
\usepackage{textcomp}
\usepackage{comment}
\usepackage[show]{ed}

\usepackage{verbatim}
\usepackage[normalem]{ulem}

\usepackage{graphicx}
\usepackage{amssymb}
\usepackage{epstopdf}
\usepackage{hyperref}
\usepackage{amsmath}
\usepackage[affil-it]{authblk}

\newcommand{\supsc}[1]{\ensuremath{^{\text{#1}}}}   

\begin{document}
\hyphenation{re-commended}

\title{Identification of functionally related enzymes by learning-to-rank methods}

\author{Michiel Stock, Thomas Fober, Eyke H\"ullermeier, Serghei Glinca, Gerhard Klebe,\newline
        Tapio Pahikkala,
        Antti Airola,
        Bernard De Baets, Willem Waegeman \thanks{M. Stock, B. De Baets and W. Waegeman are with the Department
of Mathematical Modelling, Statistics and Bioinformatics, Ghent University, Coupure links 653, 9000 Ghent, Belgium, email: firstname.surname@ugent.be.
T. Fober and E. H\"ullermeier are with the Philipps-Universit\"at of Marburg, Department of Mathematics and Computer Science, Hans-Meerwein-Stra\ss e~6, D-35032 Marburg, Germany.
S. Glinca and G. Klebe are also with the Philipps-Universit\"at of Marburg, Department of Pharmacy, Marbacher Weg 6-10, D-35032 Marburg, Germany.
T. Pahikkala and A. Airola are with the Department of Information Technology and the Turku Centre for Computer Science, University of Turku, Joukahaisenkatu 3-5 B
20520 Turku, Finland.}
}
\date{}
\maketitle

\begin{abstract}

Enzyme sequences and structures are routinely used in the biological sciences as queries to search for functionally related enzymes in online databases. To this end, one usually departs from some notion of similarity, comparing two enzymes by looking for correspondences in their sequences, structures or surfaces. For a given query, the search operation results in a ranking of the enzymes in the database, from very similar to dissimilar enzymes, while information about the biological function of annotated database enzymes is ignored.   

In this work we show that rankings of that kind can be substantially improved by applying kernel-based learning algorithms. This approach enables the detection of statistical dependencies between similarities {of the active cleft} and the biological function of annotated enzymes. This is in contrast to search-based approaches, which do not take annotated training data into account. {Similarity measures based on the active cleft are known to outperform sequence-based or structure-based measures under certain conditions.} We consider the Enzyme Commission (EC) classification hierarchy for obtaining annotated enzymes during the training phase. The results of a set of sizeable experiments indicate a consistent and significant improvement for a set of similarity measures that exploit information about small cavities in the surface of enzymes.

\end{abstract}

\section{Introduction}



Modern high-throughput technologies in molecular biology are generating more and more protein sequences and tertiary structures, of which only a small fraction can ever be experimentally annotated w.r.t.\ functionality. Predicting the biological function of enzymes\footnote{Enzymes are biomolecules that catalyze chemical reactions. All the enzymes we consider in this work are proteins and vice versa. Both notions will be used interchangeably.} remains extremely challenging, and especially novel functions are hard to detect, despite the large number of automated annotation methods that have been introduced in the last decade~\cite{Friedberg2006, Dunaway-Mariano2008}. 


Existing online services, such as BLAST or ReliBase, often provide tools to search in databases that contain collections of annotated enzymes. These systems rely on some notion of similarity when searching for related enzymes, but the definition of similarity differs from system to system. Indeed, a vast number of measures for expressing similarity between two enzymes exists in the literature, performing calculations on different levels of abstraction. One can make a major subdivision of these measures into approaches that solely use the sequence of amino acids, approaches that also take into account the tertiary structure, and approaches that consider local fold information by analyzing small cavities (hypothetical binding sites) at the surface of an enzyme.

Sequence-based measures such as BLAST~\cite{Altschul1990} and PSI-BLAST~\cite{Altschul1997} can be computed in an efficient manner and are able to find enzymes with related functions under certain conditions. In addition to these, several kernel-based methods have been developed to make predictions for proteins at the sequence level - see e.g.~\cite{Leslie2002, Leslie2004}.
A high sequence similarity usually results in a high structural similarity, and
proteins with a sequence identity (the number of matches in an alignment) above 40 \% are generally considered to share the same structure~\cite{Powers2006}. However, this assumption becomes less reliable in the twilight zone, when the sequence identity is situated between 25 and 40 \%. Furthermore, enzymes with comparable functions can exhibit sequences with very low sequence identity~\cite{Rost2002,Chalk2004}. 

For these reasons, and because three-dimensional crystal structures are becoming more and more available in online databases, the comparison of proteins at the structural level has gained increasing attention. The secondary structure of an enzyme is known to highly influence its biological function~\cite{Kinoshita2007} and contains valuable information that is missing at the sequence level~\cite{Thornton2000, Kotera2004a,Egelhofer2010}. Many approaches that perform calculations on the overall fold of the protein have been developed - see e.g.\ ~\cite{Shatsky2004, Shatsky2006a, Harrison2003}. Unfortunately, such approaches are also not optimal for determining the function of enzymes. They require knowledge of active site residues and usually lead to a quite coarse representation, especially for enzymes, where often only a few specific residues are responsible for the catalytic mechanism~\cite{Martin1998}. %
For example, the vicinal-oxygen-chelate superfamily shows a large functional diversity while having only a limited sequence diversity~\cite{Babbitt1997,Gerlt2001}.
It has also been shown that some parts of the protein structure space have a high functional diversity~\cite{Osadchy2011}, further limiting the use of global fold similarity. For these reasons, many methods consider local structural features, such as evolutionary conserved residues~\cite{Kristensen2008, Erdin2011}.

The most appropriate similarity measures for the prediction of enzyme functions focus on surface regions in which ligands, co-factors and substrates can bind~\cite{1996Laskowski}. Cavities in the surface are known to contain valuable information, and exploiting similarities between those cavities helps finding functionally related enzymes. By considering structural and physico-chemical information of binding sites, one can detect relationships that cannot be found using traditional sequence- and fold-based methods, making such similarities of particular interest for applications in drug discovery~\cite{Perot10,Andersson2010}. In addition to providing a complementary notion of protein families~\cite{Weisel2010}, these methods also allow for extracting relationships between cavities of unrelated proteins~\cite{Weber2004}. Similarity measures that highlight cavities and binding sites can be further subdivided into graph-based approaches such as~\cite{Huan1999,Borgwardt2005,Gartner2008,Shervashidze2009, Vacic2010}, geometric approaches such as~\cite{Shatsky2006,Fober2011} and feature-based approaches such as~\cite{Weill2010,Fober2012}. These measures will be discussed more thoroughly in Section~2. 

This paper aims to show that the search for functionally related enzymes can be substantially improved by applying \emph{learning-to-rank} methods. {These algorithms use training data to build a mathematical model for ranking objects, such as enzymes, that are not necessarily seen among the training data.} While our methods can be applied to all types of data, as long as a meaningful similarity measure can be constructed, we only demonstrate its power using cavity-based measures, for the reasons explained above. Ranking-based machine learning algorithms are often used for applications in information retrieval~\cite{Hullermeier2010a}.  Due to their proven added value for search engines, ranking-based machine learning methods have gained some popularity in bioinformatics, for example in drug discovery \cite{Rathke:2010cz,Agarwal2010a} or to find similarities between proteins~\cite{Weston2004,Kuang2005}. 
Despite this, many online services such as BLAST, PDB, Dali and CavBase solely rely on similarity measures to construct rankings, without utilizing annotated enzymes and learning algorithms to steer the search process during a training phase. However, due to the presence of annotated enzymes in online databases, improvements can be made by applying ranking-based machine learning algorithms. This amounts to a transition from an unsupervised to a supervised learning scenario.

Using four different cavity-based similarity measures and one based on sequence alignment as input for RankRLS~\cite{Pahikkala2009}, a kernel-based ranking algorithm, we demonstrate a significant improvement for {each of these measures}. RankRLS works in a similar way as competitors such as RankSVM~\cite{Joachims2005}, because it uses annotated training data to learn rankings during a training phase. The training data is annotated via the Enzyme Commission (EC) functional classification hierarchy, a commonly used way to subdivide enzymes into functional classes. EC numbers adopt a four-number hierarchical structure, representing different levels of catalytic detail. Importantly, this representation focuses on the chemical reactions that are performed, and not on structure or homology.
As explained more elaborately in Section~2, the EC numbers are used to construct a ground-truth catalytic similarity measure, and subsequently to generate ground-truth rankings. In addition to obtaining annotated training data, this procedure also allows for a fair comparison with the more traditional approach, using conventional performance measures for rankings. This way of evaluating also characterizes the difference between our search engine approach and previous work in which supervised learning algorithms for EC number assignment have been considered -- for a far from complete list see e.g.\ ~\cite{Dobson2005a,Borgwardt2005,Rousu2006,Sokolov2008,Arakaki2009}. In this work we are unable to compare to such methods, because they do not return rankings as output. Nonetheless, similar to some of these approaches, we do take the hierarchical structure of the EC numbers into account. {Instead of predicting one EC number, a ranking of functionally related enzymes is returned for a given query}. In this scheme the top of the obtained ranking is expected to contain enzymes with functions similar to the query enzyme with an unknown EC number. A ranking provides end users with a generally well-known and easily understandable output, while still useful results can be retrieved when an enzyme with a new EC number is encountered.  

\section{Material and methods}\label{MatandMet}

\subsection{Database}\label{database}
Our work builds upon CavBase, a database that is made commercially available as part of ReliBase~\cite{Hendlich2003}. CavBase can be used for the automated detection, extraction, and storage of protein cavities from experimentally determined protein structures, which are available through the Protein Data Bank (PDB). 

The geometrical arrangement of the pocket and its physico-chemical properties are first represented by predefined pseudocenters -- spatial points that characterize the geometric center of a {functional group specified by a} particular property. The type and the spatial position of the pseudocenters depend on the amino acids that border the binding pocket and expose their functional groups. They are derived from the protein structure using a set of predefined rules~\cite{Schmitt2002}. Hydrogen-bond donor, acceptor, mixed donor/acceptor, hydrophobic aliphatic, metal ion, pi (accounts for the ability to form $\pi$-$\pi$ interactions) and aromatic properties are considered as possible types of pseudocenters. These pseudocenters can be regarded as a compressed representation of surface areas where certain protein-ligand interactions are encountered. Consequently, a set of pseudocenters is an approximate representation of a spatial distribution of physico-chemical properties.

To build and test our models we require an appropriate data set that contains sufficiently many proteins and EC classes. Based on the experience of local pharmaceutical experts, we chose the data set of EC classes depicted in Table \ref{enzlist}. To generate the first data set (data set I), we retrieved all proteins from the PDB which got assigned one of these EC classes. Thus, we ended up with a set of 5,257 proteins. To ensure that only unique proteins were contained in our data set, we used the protein culling server\footnote{http://www.bioinf.manchester.ac.uk/leaf/} with its default parameterization. As such, all proteins that have high pairwise homology were filtered out. This procedure resulted in a data set of cardinality 1,714. To extract the active site of the protein we used the assumption that the largest binding site of a protein does contain its catalytic center \cite{1996Laskowski}. Hence, for each protein we took the binding site from the database CavBase which maximized the volume. From our data set, 158 proteins were not contained in the CavBase (e.g., because the structure was determined by NMR instead of X-ray). Therefore these proteins were removed from the data set, resulting in a final data set of size 1,556.

The first data set comes with two drawbacks. First of all, the binding site containing the catalytic centre was determined by a pure heuristic, namely by taking the largest binding site among all binding sites a protein exhibits. Moreover, sufficient resolution was not a criterium for selecting the cavities. This may lead to a data set of low quality. Therefore, relying on the expertise of pharmaceutical experts we compiled another data set referred to as data set II, containing the same EC classes. For this data set, all proteins from the PDB that have a resolution of at least 2.5 \AA{} were considered. Moreover the binding site volume was required to range between 350 \AA{}\supsc{3} and 3500 \AA{}\supsc{3}. Structures not meeting these conditions were eliminated since resolutions below 2.5 \AA{} usually lead to a too coarse representation, while binding sites with volumes outside the above-mentioned range are usually artefacts produced by the algorithm used for their detection. From the resulting set of 24,102 proteins the active site was selected. This resulted in a data set of 1730 enzymes on which we applied the protein culling server to finally end up with a second data set of 561 enzymes.

A pairwise sequence similarity matrix and phylogenetic tree of our data sets can be found in the supplementary materials.

\begin{table}[h] 
\begin{center}
\caption{List of the 21 EC numbers with their accepted name and the number of examples of each class for the two data sets.}\label{EC_list}

\resizebox{!}{0.26\textwidth}{

\begin{tabular}{llcc}

\hline
 EC number & accepted name & \# set I  & \# set II \\
 \hline 
 EC 1.1.1.1 & alcohol dehydrogenase & 23 & 15 \\ 
EC 1.1.1.21 & aldehyde reductase & 35 & 30 \\ 
EC 1.5.1.3 &  dihydrofolate reductase & 110 & 6 \\ 
EC 1.11.1.5 & cytochrome-c peroxidase & 92 & 31 \\ 
EC 1.14.15.1 & camphor 5-monooxygenase & 30 & 36 \\ 

EC 2.1.1.45 & thymidylate synthase & 63 & 22 \\ 
EC 2.1.1.98 & diphthine synthase & 5 & 43 \\ 
EC 2.4.1.1 & phosphorylase & 43 & 40 \\ 
EC 2.4.2.29 & tRNA-guanine transglycosylase & 32 & 16 \\ 
EC 2.7.11.1 & non-specific serine/threonine enzyme kinase & 304 & 24 \\

EC 3.1.1.7 & acetylcholinesterase & 23 & 13 \\ 
EC 3.1.3.48 & enzyme-tyrosine-phosphatase & 151 & 28 \\ 
EC 3.4.21.4 & trypsin & 118 & 72 \\ 
EC 3.4.21.5 & thrombin & 87 & 51 \\ 
EC 3.5.2.6 & $\beta$-lactamase & 153 & 8 \\ 

EC 4.1.2.13 & fructose-bisphosphate aldolase & 48 & 4 \\ 
EC 4.2.1.1 & carbonate dehydratase & 186 & 76 \\ 
EC 4.2.1.20 & tryptophan synthase & 13 & 7 \\ 

EC 5.3.1.5 & xylose isomerase & 18 & 21 \\ 
EC 5.3.3.1 & steroid $\Delta$-isomerase & 14 & 10 \\ 

EC 6.3.2.1 & pantoate-$\beta$-alanine ligase & 8 & 8 \\
 \hline
 \end{tabular}}
 \label{enzlist}
 \end{center}
 \end{table}

\subsection{Similarity measures for cavities}\label{SME}

In the introduction we have motivated why our analysis is restricted to similarity measures for cavities, which are three-dimensional objects that can be represented in multiple ways. Some measures are graph-based, transforming cavities into node-labeled and edge-weighted graphs. This allows to apply traditional techniques to compare graphs (e.g.~\cite{Huan1999}). Unfortunately, techniques that construct a boolean {similarity measure}, such as those based on graph isomorphisms, are not appropriate for comparing noisy and flexible protein structures. Computing the maximum common subgraph~\cite{Bunke1998} can be considered as a more appropriate alternative, and this method will be used in this paper as a baseline (see below). The graph edit distance~\cite{Sanfeliu1983} is another measure to compare graphs, specifying the number of edit operations needed to transform a given graph into another graph. This distance can be calculated in different ways, e.g.,~by using a greedy heuristic~\cite{Weskamp2007} or quadratic programming~\cite{Neuhaus2007}. Unfortunately, {the graph edit distance} is very hard to parameterize and often quite inefficient. More efficient approaches belong to the class of graph kernels. They have gained a lot of attention in bioinformatics, as they allow for a sufficiently high degree of error tolerance. Different realizations are available, such as the shortest path kernel~\cite{Borgwardt2005a}, the random walk kernel~\cite{Gartner2008} and the graphlet kernel~\cite{Shervashidze2009, Vacic2010}. Graph kernels work particularly well for small molecules such as ligands, but they are less useful for larger molecules such as proteins. They gave rather poor results in~\cite{Fober2012}, which explains why we concentrated here on the maximum common subgraph as a representative for graph-based approaches. 

{As a second category of measures for cavities, geometric methods directly process the labeled spatial coordinates of the functional parts, {denoted as point clouds,} instead of transforming a protein cavity into a graph. Remarkably, only few approaches have been proposed that build on this representation.} In~\cite{Shatsky2006} geometric hashing is employed to calculate a superposition of protein cavities that can be used to derive an alignment and a similarity score. A similar approach was used in~\cite{Fober2011}, in which an optimization problem was solved instead of applying geometric hashing. Beside these two approaches, several other methods exist for comparing two point clouds~\cite{Alt1996}. Unfortunately, the majority of these methods cannot cope with biological data, due to a very high complexity or error intolerance. 

As a third family of approaches, one can also represent the protein cavity as a feature vector, taking both the geometry of the cavity and physico-chemical properties into account -- see e.g.\ ~\cite{Fober2012, Weill2010}. Subsequently, traditional or specialized measures can be applied on these vectors to obtain similarity scores between protein cavities~\cite{Mahe2005,Deza2009}.

In the experiments we selected representative method for each of the three groups: one graph-based measure, one geometric measure and one feature-based measure. We also considered the original CavBase measure and a measure obtained from the Smith-Waterman protein sequence alignment. This lead to a comparison of five different measures, four based on cavities and one based on sequence alignment. Below, these measures are explained more in detail: 
\begin{itemize}

\item[\textbf{Labeled Point Cloud Superposition (LPCS)}]~\cite{Fober2011}. This value is obtained by processing labeled point clouds. Hence, the CavBase data can be used directly without a need for transforming it into another representation. Intuitively, two labeled point clouds are considered similar if they can be spatially superimposed. More specifically, an approximate superposition of the two structures is obtained by fixing the first point cloud and �moving� the second point cloud as a whole. Two point clouds are well superimposed when each point in the first cloud can be matched with a point in the second point cloud, while the distances of these points are small and their labels consistent. This concept is used to define a fitness function that is maximized using a direct search approach~\cite{Beyer2002}. The obtained maximal fitness is taken as the similarity between the two labeled point clouds. A similar measure was also proposed in~\cite{Hoffmann2010}, but a convolution kernel is suggested to obtain similarities between the point clouds.  

\item[\textbf{Maximum Common Subgraph (MCS)}]~\cite{Bunke1998}. Using the {MCS}, the original representation in the form of a labeled point cloud must be transformed into a node-labeled and edge-weighted graph. Each pseudocenter is becoming a node labeled with the corresponding physico-chemical property. To capture the geometry, a complete graph is considered, where each edge is weighted with the Euclidean distance between the two pseudocenters it is adjacent to. The problem of measuring similarity between protein cavities now boils down to the problem of measuring similarity between graphs. A well-known approach here is to search for the maximum common subgraph of the two input graphs and to define similarity as the size of the maximum common subgraph relative to the size of the larger graph. In case of noisy data, a threshold $\epsilon$ is required, defining two edges as equal if their weight differs at most by $\epsilon$. In this paper, this parameter is set to 0.2 \AA, as recommended by several authors \cite{Weskamp2007,Fober2009}.

\item[\textbf{CavBase (CB) similarity}]~\cite{Schmitt2002}. CavBase also makes  use of an algorithm for the detection of common subgraphs. Instead of considering the largest common subgraph, as done in the case of MCS, the 100 largest common subgraphs are considered. Each common subgraph is used to {determine} a transformation rule by means of the Kabsch algorithm~\cite{Kabsch1976}, which superimposes both proteins. In a post-processing step the surface points are also superimposed according to the transformation rule, and a similarity score is derived using these surface points. Eventually, a set of 100 similarity values is obtained, from which the highest value is returned as similarity between the two protein cavities.

\item[\textbf{Fingerprints (FP)}] Fingerprints are a well-known concept and have been used successfully in many domains. For the comparison of protein binding sites, the authors in \cite{Fober2009b} transformed the protein binding site into a node-labeled and edge-weighted graph as described above. Moreover they defined generically a set of features, namely complete node-labeled and edge-weighted graphs of size 3. For each such feature, a test is performed to decide whether or not the feature is contained in the graph representing the protein. This is done by subgraph isomorphism, to checks whether the labels are identical. The nodes of the features are labeled by the set of physiochemical properties. Edges of patterns are labeled by intervals or bins and instead of testing for equivalence, a test is performed whether edge weight of the graph representing the protein falls into the bin of the pattern. The thus generated fingerprints are compared by means of the Jaccard similarity measure, as proposed by~\cite{Mahe2005}. 

\item[\textbf{Smith-Waterman (SW)}] Beside using structure-based approaches to compare protein binding sites, we used also sequence alignment in our experimental study. To calculate sequence alignments we used the Smith-Waterman algorithm \cite{Smith1981} which was parameterized with the Blosum-62 matrix. From the sequence alignment we derived the sequence identity which was subsequently used to perform experiments.

\end{itemize}

\subsection{Unsupervised ranking}\label{unsupervised}
In the introduction we have explained why existing online services such as BLAST, PDB, Dali and CavBase construct rankings in an unsupervised way. These systems create a ranking by means of a similarity measure only, without training a model that uses annotated enzymes. 
The annotated enzymes in a database are simply ranked according to their similarity with an enzyme query with unknown function. In the case of CavBase, the enzymes having a high cavity-based similarity appear on top of the ranking, those exhibiting a low cavity-based similarity end up at the bottom. More formally, let us represent the similarity between a pair of enzymes by $K : \mathcal{V}^2  \rightarrow \mathbb{R}^+$, where $\mathcal{V}$ represents the set of all potential enzymes. Given the similarities $K(v,v')$ and $K(v,v'')$, we compose the ranking of $v'$ and $v''$ conditioned on the query $v$ as:
\begin{equation}
\label{eq:kernelranking}
v'  \succeq_v^K v'' \Leftrightarrow K(v,v') \geq K(v,v'')\,,
\end{equation}
{where $\succeq_v^K$ indicates the relation \emph{is ranked higher than, for query $v$, based on the similarity $K$}}. Note that this is a relation between two enzymes \emph{conditioned} on a third enzyme. In our context, there is no meaningful ranking possible between enzymes $v'$ and $v''$ without referring to another enzyme $v$.
 This approach adopts the same methodology as a nearest neighbor classifier, but a ranking rather than a class label should be seen as the output of the algorithm.  

The quality of such rankings can be evaluated when the database contains annotated enzymes and annotated queries. In an evaluation phase, we compare the obtained ranking with the \textit{ground truth} ranking, which can be constructed from the EC numbers for annotated enzymes. This ground truth ranking can be deduced from the catalytic similarity (i.e., ground truth similarity) between the query and all database enzymes, by counting the number of successive matches in the EC label of the query and the database enzymes. Thus the catalytic similarity is a property of only a pair of enzymes. In contrast, in order to create the ground truth ranking of two enzymes, the catalytic similarity has to be calculated w.r.t.\ a third enzyme. For example, an enzyme with EC number EC~2.4.2.23 has a catalytic similarity of two compared to an enzyme labeled as EC~2.4.99.12, since both enzymes belong to the family of glycosyltransferases. Conversely, the same enzyme manifests a similarity value of only one with an enzyme labeled as EC~2.8.2.23. Both are transferases in this case, but they show no further relevant similarity in the chemistry of the reactions to be catalyzed. 
\begin{figure}[t]
\centering
\includegraphics[width=0.35\textwidth]{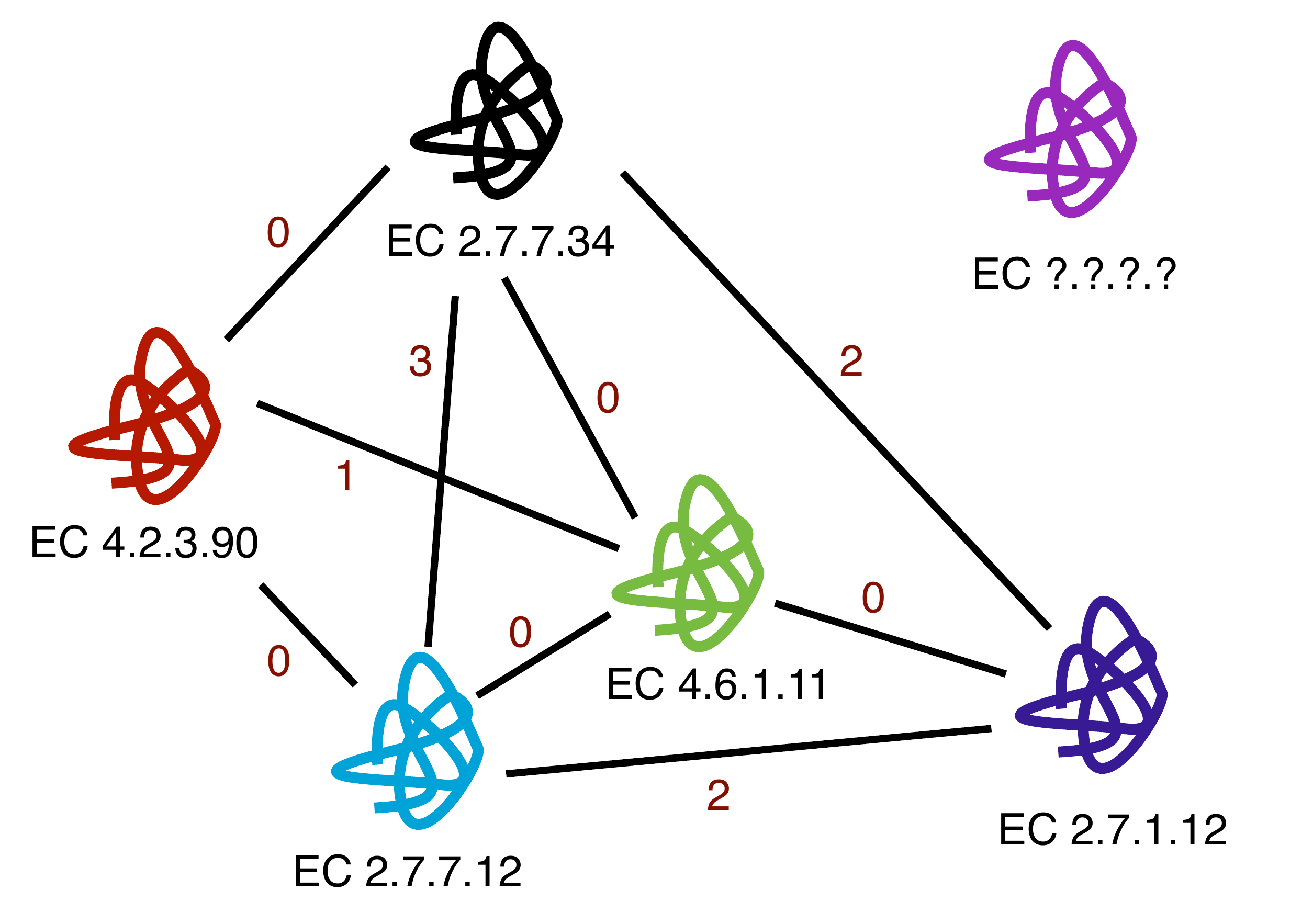}
\caption{Six enzyme structures are shown, five of which correspond to a known EC number. The catalytic similarity $Q$ is depicted on the edges of the graph. The algorithm that we present allows us to infer for the unannotated query (denoted as EC ?.?.?.?) a ranking of the annotated enzymes. To this end, the unsupervised approach solely uses cavity-based similarity measures, whereas the supervised approach also takes the EC numbers of annotated enzymes into account.}\label{ECsim}
\end{figure}

More formally, let us represent the catalytic similarity between two enzymes by a relation $Q : \mathcal{V}^2  \rightarrow \{0, 1,2, 3,4\}$. $Q$ is defined by:
\begin{equation}
Q(v,v') = \sum^4_{i=1}\prod_{j=1}^iq_iq_j\,,\nonumber
\end{equation}
where $q_i$ equals 1 if the $i$th digit of the EC numbers of $v$ and $v'$ are the same and 0 otherwise. Figure~\ref{ECsim} gives an example for six enzymes, five of which correspond to a known EC number.
The catalytic similarity $Q$ is depicted on the edges of the graph. The proposed algorithm allows us to infer for an unannotated query a ranking of the annotated enzymes, {some of which the algorithm may not have encountered among the training data.}

Given the similarities $Q(v,v')$ and $Q(v,v'')$, we compose similar to Eq.~(\ref{eq:kernelranking}) the ground truth ranking of $v'$ and $v''$  conditioned on the query $v$ as:
\begin{equation}
\label{eq:gtranking}
v'  \succeq_v^Q v'' \Leftrightarrow Q(v,v') \geq Q(v,v'')\,.\nonumber
\end{equation}
As a result, an entire ground truth ranking of database enzymes with known EC numbers can be constructed, given an annotated query enzyme. 

\subsection{Supervised ranking}\label{SRE}
In contrast to unsupervised ranking approaches, supervised algorithms do take ground truth information into account during a training phase. We perform experiments with so-called conditional ranking algorithms~\cite{Pahikkala2010, pahikkala2013conditional} using the RankRLS implementation~\cite{Pahikkala2009}. Let us introduce the short-hand notation $e=(v,v')$ to denote a couple consisting of an enzyme query $v$ and a database enzyme $v'$. RankRLS produces a linear basis function model of the type: 
\begin{equation}
h(e)= h(v,v') = \langle\mathbf{w},\Phi (e)\rangle\,, \label{linmod}
\end{equation}
in which $\mathbf{w}$ denotes a vector of parameters and $\Phi(e)$ an implicit feature representation for the couple $e=(v,v')$. 

RankRLS differs from more conventional kernel-based methods because it optimizes a convex and differentiable approximation of the rank loss in bipartite ranking (i.e., area under the ROC curve) instead of the zero-one loss. Together with the standard L2 regularization term on the parameter vector $\mathbf{w}$ and a regularization parameter $\lambda$, the following loss is minimized:
\begin{equation}
\mathcal{L}(h,T)=\sum_{v\in V} \sum_{e,\bar{e}\in E_v}(Q_e-Q_{\bar{e}}-h(e)+h(\bar{e}))^2\,, \label{CDRE}
\end{equation}
for a given training set $T=\{(e,Q_e) \mid e\in E\}$. Here $Q_e = Q(v,v')$ denotes the ground truth similarity as defined above, $E$ the set of training query-object couples for which ground truth information is available and $E_v$ the subset of $E$ containing results for the query $v$. The outer sum in Eq.~(\ref{CDRE}) takes all queries into account, and the inner sum analyzes all pairwise differences between the ranked results for a given query. This loss can be minimized in a computationally efficient manner, using analytic shortcuts and gradient-based methods, as shown in~\cite{Pahikkala2010,pahikkala2013conditional}. 

According to the representer theorem~\cite{Scholkopf2002}, one can rewrite Eq.~(\ref{linmod}) in the following dual form:
\begin{equation}
h(e)=\langle\mathbf{w},\Phi (e)\rangle=\sum_{\bar{e} \in E}a_{\bar{e}} K^\Phi(e, \bar{e})\,.\nonumber
\end{equation}
with $K^\Phi(e, \bar{e})$ a kernel function with four enzymes as input and $a_{\bar{e}}$ the weights in the dual space. In this paper we adopt the Kronecker product feature mapping, containing information on couples of enzymes:
\begin{equation}
\Phi (e) = \Phi(v,v') = \phi(v) \otimes \phi(v')\,,\nonumber
\end{equation}
with $\phi(v)$ a feature mapping of an individual enzyme and $\otimes$ the Kronecker product. One can easily show that this pairwise feature mapping yields the Kronecker product pairwise kernel in the dual representation:
\begin{equation}
K^\Phi(e, \bar{e}) = K^\Phi(v,v ', \bar{v},\bar{v} ')=K^\phi(v, \bar{v}) K^\phi(v', \bar{v}')\,,\nonumber
\end{equation}
with $K^\phi$ a traditional kernel for enzymes. Specifying a universal kernel for $K^\phi$ leads to a universal kernel for $K^\Phi$~\cite{Waegeman2012}, {indicating that one can use the kernel to represent any arbitrary relation, provided that the learning algorithm has access to training data of sufficient quality}. This kernel has been introduced in~\cite{Ben-Hur2005} for modelling protein-protein interactions. 
We consider this kernel because of its universal approximation property, but also other pairwise kernels exist, such as the cartesian pairwise kernel~\cite{kashima2009}, the metric learning pairwise kernel~\cite{Vert2007} and the transitive pairwise kernel~\cite{Herbrich2000,Pahikkala2010}. Nonetheless, it is probably not very surprising that such kernels only yield an improvement if the concepts to be learned satisfy the restrictions that are imposed by the kernels~\cite{Waegeman2012}. 

With the exception of the FP measure, none of the similarity measures discussed in Section~\ref{SME} are strictly speaking valid kernels. Using the above construction, all similarity measures can be converted into kernels of type $K^\phi$, when they are made symmetric and positive definite. These attributes guarantee a numerically stable and unique solution of the learning algorithm. We simply enforced symmetry by averaging the similarity matrix with its transpose. Subsequently, we made the different similarity matrices positive definite by performing an eigenvalue decomposition and setting all eigenvalues smaller than $10^{-10}$ equal to zero. This method leads to a negligible loss of information compared to the numerical accuracy of our algorithms and data storage. Finally, each kernel matrix was normalized so that all diagonal elements have a value equal to one. Since these procedures were performed on the whole data set, one arrives at a so-called transductive learning setting~\cite{Chapelle2006}. Minor adjustments would obtain a more traditional inductive learning setting. Note that overfitting is prevented when applying this procedure, since the EC numbers of the enzymes in the data set are not taken into account.

Since the catalytic similarity is a symmetric measure we also perform a post-processing to the output of our algorithm. The matrix with the predicted values used for ranking the enzymes is made symmetric by averaging it with its transpose.

\subsection{Performance measures for ranking}\label{perf_meas}

The ranking obtained with unsupervised or supervised learning algorithms can be compared to the ground truth ranking by applying performance measures that are commonly used in information retrieval. 

First of all, the ranking accuracy (RA) is considered, and it is defined as follows:

\footnotesize
\begin{equation} \label{RE}
\text{RA}= \frac{1}{\lvert V \lvert  }\sum_{v\in V}  \frac{1}{\lvert \{E_v \mid y_e>y_{\bar{e}} \}\lvert}  \sum_{\substack{e,\bar{e}\in E_v : \\ y_e<y_{\bar{e}}}}I(h(e)-h(\bar{e}))\,,\nonumber
\end{equation}
\normalsize
with $I(x)$ the Heaviside function, returning one when its argument is positive, zero when its argument is negative and $1/2$ when its argument is zero. The ranking accuracy can be considered as a generalization of the area under the ROC curve for more than two ordered classes~\cite{Waegeman2008a}. 

Our interest in the ranking accuracy is motivated by two reasons. Firstly, unlike most other performance measures we consider, all levels of the EC hierarchy are taken into account to determine the performance of different algorithms. Predicted rankings can be interpreted as layered or multipartite rankings -- see e.g.\ cite{Waegeman2008e,Furnkranz2009}. The ranking accuracy preserves this hierarchical structure by counting all pairwise comparisons. The second reason of interest is based on the fact that the ranking accuracy is optimized by the RankRLS software, using the convex and differentiable approximation given in Eq.~(\ref{CDRE}). This loss function characterizes the most important difference with more traditional kernel-based algorithms, such as support vector machines, resulting in an information retrieval setting instead of a more traditional classification or network inference setting. 

Since the ranking accuracy is not generally known in bioinformatics, we also evaluated our algorithms using three more conventional performance measures that are commonly considered for bipartite rankings (i.e., rankings containing relevant versus irrelevant objects). These three measures are the area under the ROC curve (AUC), mean average precision (MAP) and normalized discounted cumulative gain (nDCG)~\cite{Jarvelin2002}. For AUC and MAP all ground truth rankings had to be converted into bipartite rankings, leading to a decrease in granularity for performance estimation. We chose a cut-off threshold of three in ground truth similarity: a retrieved enzyme is relevant to the enzyme query if at least the first three parts of its EC number are identical to the query.

\subsection{Experimental setup}

We selected two data sets of enzymes from CavBase, as described in Section~\ref{database}. The ground-truth catalytic similarity of all enzyme pairs was computed for each data set. Each data set was further randomized and split in four cross-validation folds of equal size. In the unsupervised case each subset was used individually to allow for a comparison with the supervised model. Of such a subset, each enzyme was used as a query to rank the remaining enzymes, as described in Section~\ref{unsupervised}. The performance for each of these rankings was averaged to obtain the global performance over the folds.

In the supervised setting, each fold was withheld as a test set, while the other three parts of the data set were used for training and model selection. This process was repeated for each part, so that every instance was used for training and testing (thus, four-fold outer cross-validation). {Neither the query nor the database enzymes are thus used for building a model, which allows us to demonstrate that our methods can generalize to new enzymes.} In addition, a 10-fold inner cross-validation loop was implemented for estimating the optimal regularization parameter $\lambda$, as recommended in~\cite{Varma2006}. The value of this hyperparameter, which controls model complexity, was selected from a grid containing all powers of 10 from $10^{-4}$ to $10^5$.
 The final model was trained using the whole training set and the median of the best hyperparameter values over the ten folds. We used the implementation RLScore\footnote{See http://staff.cs.utu.fi/\texttildelow aatapa/software/RLScore/ for this software.} in Python to train the models. 

\section{Results and discussion}

\begin{table*} 
\begin{center}
\caption{Summary of the results obtained for unsupervised and supervised ranking for both data sets. For each combination of similarity and type of performance measure, the performance is averaged over the different folds and queries, with the standard deviation between parentheses.
}\label{results_unsup}
\begin{footnotesize}
\begin{tabular}{c*{6}{l}}
\hline
\hline
&\multicolumn{6}{c}{Set I}\\
\hline
& & CB & FP  & MCS & LPCS & SW\\
 \hline 
\multirow{4}{*}{Unsupervised} & RA & 0.6096 (0.1286) &0.6411 (0.1670) & 0.5960 (0.1010) & 0.6173 (0.1244) & 0.6576 (0.1446) \\
& MAP & 0.4883 (0.2912) & 0.4478 (0.2692) & 0.4967 (0.2659) & 0.4826 (0.2572) & 0.5201 (0.2964)\\
& AUC & 0.7110 (0.1845) & 0.7221 (0.1961) & 0.6967 (0.1728) & 0.7183 (0.1565) & 0.7397 (0.1980)\\
& nDCG & 0.7161 (0.3107) & 0.6615 (0.2709) & 0.7686 (0.3001) & 0.7364 (0.276) & 0.7401 (0.2597)\\

 \hline 
\multirow{4}{*}{Supervised} & RA & 0.7717 (0.1960) & 0.7175 (0.2098) & 0.7988 (0.1906) & 0.7741 (0.1945) & 0.7899 (0.1789)\\
& MAP & 0.5659 (0.3384) & 0.5966 (0.3219) & 0.6854 (0.3001) & 0.6324 (0.3064) & 0.6963 (0.3103) \\
& AUC & 0.8242 (0.1857) & 0.8003 (0.2225) & 0.8585 (0.1802) & 0.8306 (0.1954) & 0.8813 (0.1591)\\
& nDCG & 0.6550 (0.4195) & 0.6928 (0.3984) & 0.7717 (0.3635) & 0.7324 (0.3724) & 0.7518 (0.3669)\\

 \hline
 \hline
 
 &\multicolumn{6}{c}{Set II}\\
\hline
& & CB & FP  & MCS & LPCS & SW\\
 \hline 
\multirow{4}{*}{Unsupervised} & RA & 0.7216 (0.1911) & 0.7212 (0.1546) & 0.8070 (0.1736) & 0.7515 (0.1647) & 0.7856 (0.1591) \\
& MAP & 0.9156 (0.1662) & 0.7156 (0.2478) & 0.9094 (0.1514) & 0.8303 (0.1768) & 0.9167 (0.1634) \\
& AUC & 0.9415 (0.1209) & 0.8714 (0.1447) & 0.9622 (0.0847) & 0.8937 (0.1165) & 0.9523 (0.1146) \\
& nDCG & 0.9599 (0.0708) & 0.8189 (0.2003) & 0.9330 (0.0948) & 0.8986 (0.1146) & 0.9589 (0.0768) \\

 \hline 
\multirow{4}{*}{Supervised} & RA & 0.9997 (0.0011) & 0.931 (0.1096) & 0.9944 (0.0253) & 0.9883 (0.0583) & 0.9985 (0.0076) \\
& MAP &  0.9995 (0.0037) & 0.905 (0.2397) & 0.9980 (0.0265) & 0.9663 (0.1172) & 0.9980 (0.0393) \\
& AUC & 0.9999 (0.0005) & 0.9844 (0.0502) & 0.9997 (0.0036) & 0.9926 (0.0354) & 0.9998 (0.0042) \\
& nDCG & 0.9983 (0.0185) & 0.8887 (0.2243) & 0.9744 (0.0773) & 0.9467 (0.1183) & 0.9968 (0.0344) \\

 \hline
 \hline
 
 \end{tabular}
 \end{footnotesize}
 \label{pal}
 \end{center}
 \end{table*}
 
 \subsection{Differences between cavity-based similarities and data sets}

Table~\ref{results_unsup} gives a global summary of the results obtained for the unsupervised and the supervised ranking approach for both data sets. One can note a sizeable difference between the performances for the different cavity-based similarities, data sets and the performance measures used. Despite this variation, it is clear that data set I is considerably harder than data set II. This can easily be explained by the fact that data set II only contains enzymes with a certain resolution of the active site. Furthermore, for set II the active site is determined by an expert, while for set I the active site is resolved by heuristically choosing the largest cavity. It is likely that some mistakes are made in this annotation process. Consequently, inferring functional similarity of data set I will be harder.

The cavity-based similarity measure based on fingerprints usually results in the worst performance, except for the ranking error in the unsupervised setting. It seems that the performance of the FP does not improve as much in the supervised approach, compared to other cavity-based similarity measures. This is likely because the fingerprints cause a high loss of information, since even functionally dissimilar enzyme cavities can be considered similar according to this metric.

Comparing the two graph-based similarities (MCS and CB), we see some differences between the data sets. Though both perform relatively well, MCS performs better for data set I, while CB is the clear champion of data set II. The good performance of CavBase for data set II can be explained easily. CavBase computes the 100 largest common subgraphs, which could be used to construct a cavity-based similarity measure. However, a graph representation leads to a loss of information, since the coordinates of pseudocenters cannot be restored. Moreover, since the size of the (maximum) common subgraph is an integer that usually lies in the range of 4 to 12 nodes, there is a loss of resolution by mapping many different pairs of cavities to the same similarity score. In theory, MCS suffers from these drawbacks. Even though the resolution problem is to a certain extent solved if the size of the maximum common subgraph is divided by the size of the larger binding site, the graph representation could still lead to a slight loss of information.

On the other hand, the LPCS measure uses geometric information, hence, no loss of information is introduced by transforming the pseudocenter representation into a graph representation. Moreover, this transformation does not cause the resolution problem. Yet, the measure is computed via solving a multimodal optimization problem, so it is possible to get stuck in a local optimum, resulting in a similarity score that is too low. Similar to MCS, LPCS seems to perform relatively better for data set I compared to data set II, probably because the local optimum becomes less of an issue in the former case. This can be explained by the fact that data set I contains larger cavities, on average, hence making it harder to find the global optimum.

Finally, we consider the measure based on sequence alignment. For data set I, the SW similarity measure competes with the MCS as one of the best measures, depending on the performance measure. For data set II in the supervised case, it is only outperformed by the CB measure. It is clear that the SW measure is a powerful method for comparing cavities, as it is also limited by a bad resolution of the cleft.. Like MCS, SW seeks to quantify the largest similar region, here as a local alignment. As this contains, information about the common residues of the cavity, this is a simple, though powerful measure.

\subsection{The benefits of supervised ranking}

The ranking of both data set I and data set II showed a considerable improvement in the supervised approach. Three important reasons can be put forward to explain the improvement in performance. In this section we will illustrate this using data set II, as this one showed the most clear effects of learning. First of all, the traditional benefit of supervised learning plays an important role. One can expect that supervised ranking methods outperform unsupervised methods, because they take annotations into account during the training phase to guide the model towards retrieval of enzymes with a similar EC number. Conversely, unsupervised methods solely rely on a meaningful similarity measure between enzymes, while ignoring EC numbers.

\begin{figure*}[t]
\centering
\begin{subfigure}{0.7\textwidth}
  \centering
  \includegraphics[width=\linewidth]{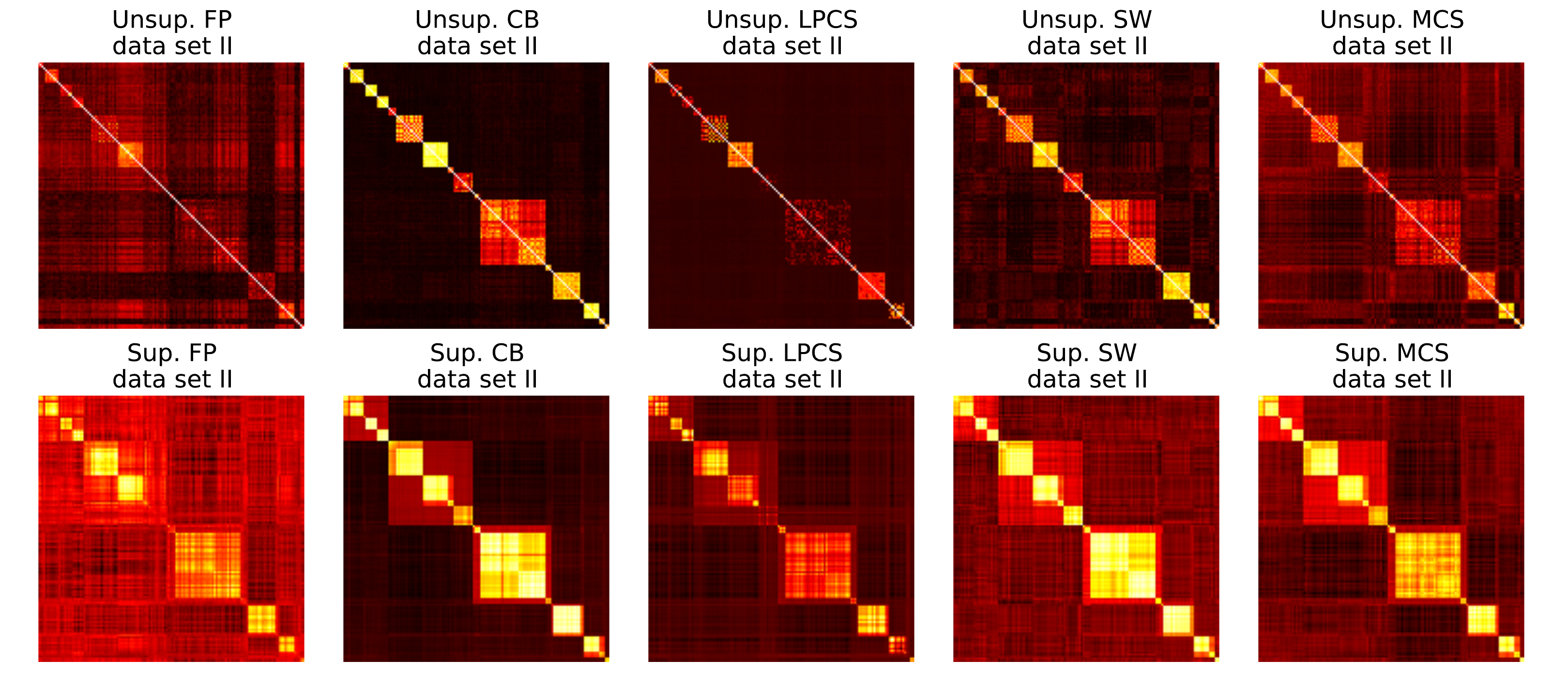}
  \caption{Unsupervised and supervised measures used for ranking}
  \label{RankVals}
\end{subfigure}%
\begin{subfigure}{.25\textwidth}
  \includegraphics[width=\linewidth]{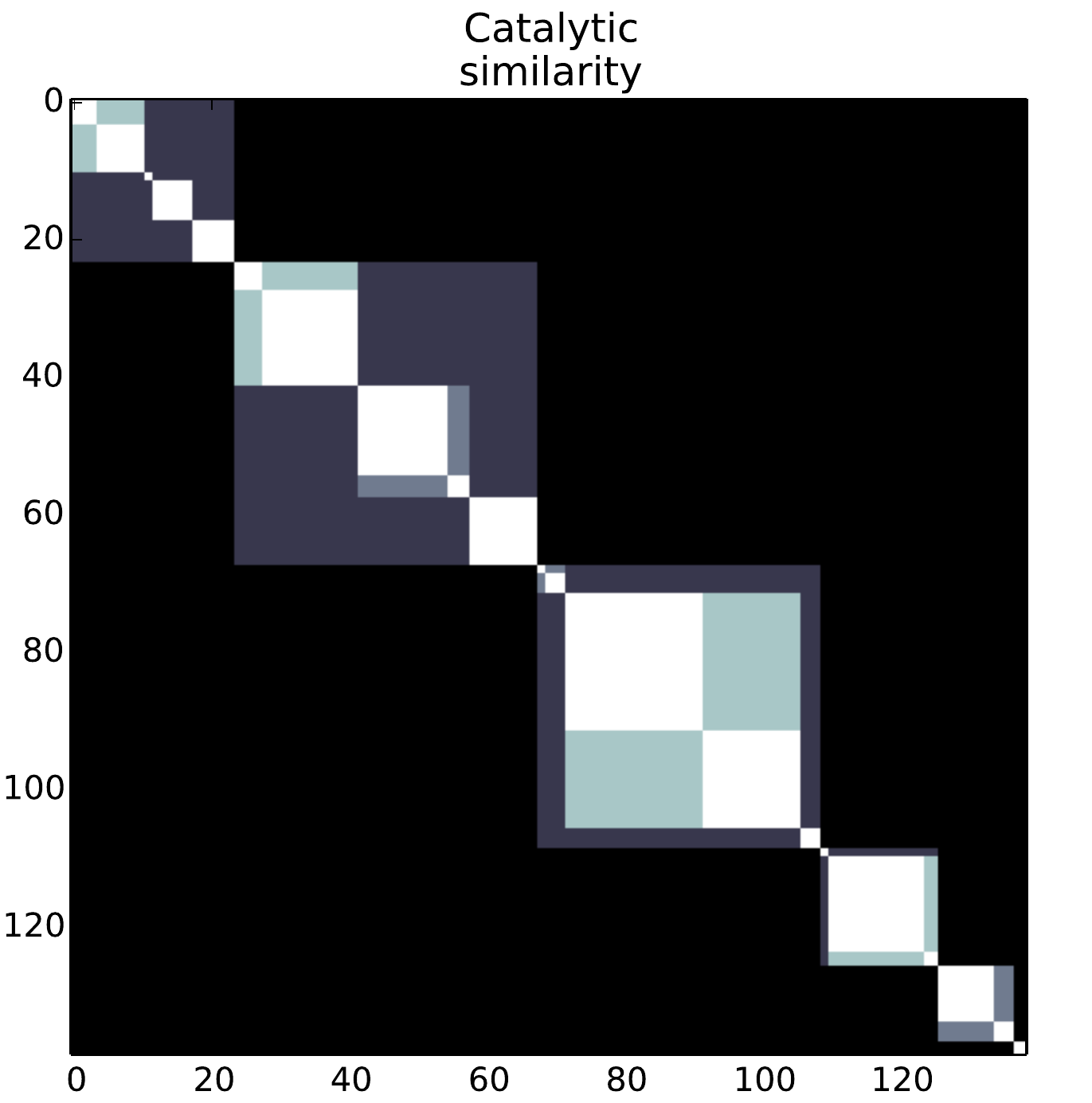}
  \caption{Ground truth}
  \label{GroundTr}
\end{subfigure}
\caption{(a) Heatmaps of the values used for ranking the data set II for one fold in the testing phase. Each row of the heatmap corresponds to one query. The four figures on top visualize the cavity-based similarities $K^\phi(v,v')$ that are used to construct an unsupervised ranking. The four figures at the bottom visualize the model output $h(v,v')$, which is used to derive a supervised ranking. (b) The ground truth catalytic similarity that had to be learned.}
\label{test_HM}
\end{figure*}

\begin{figure}
\centerline{\includegraphics[width=0.45\textwidth]{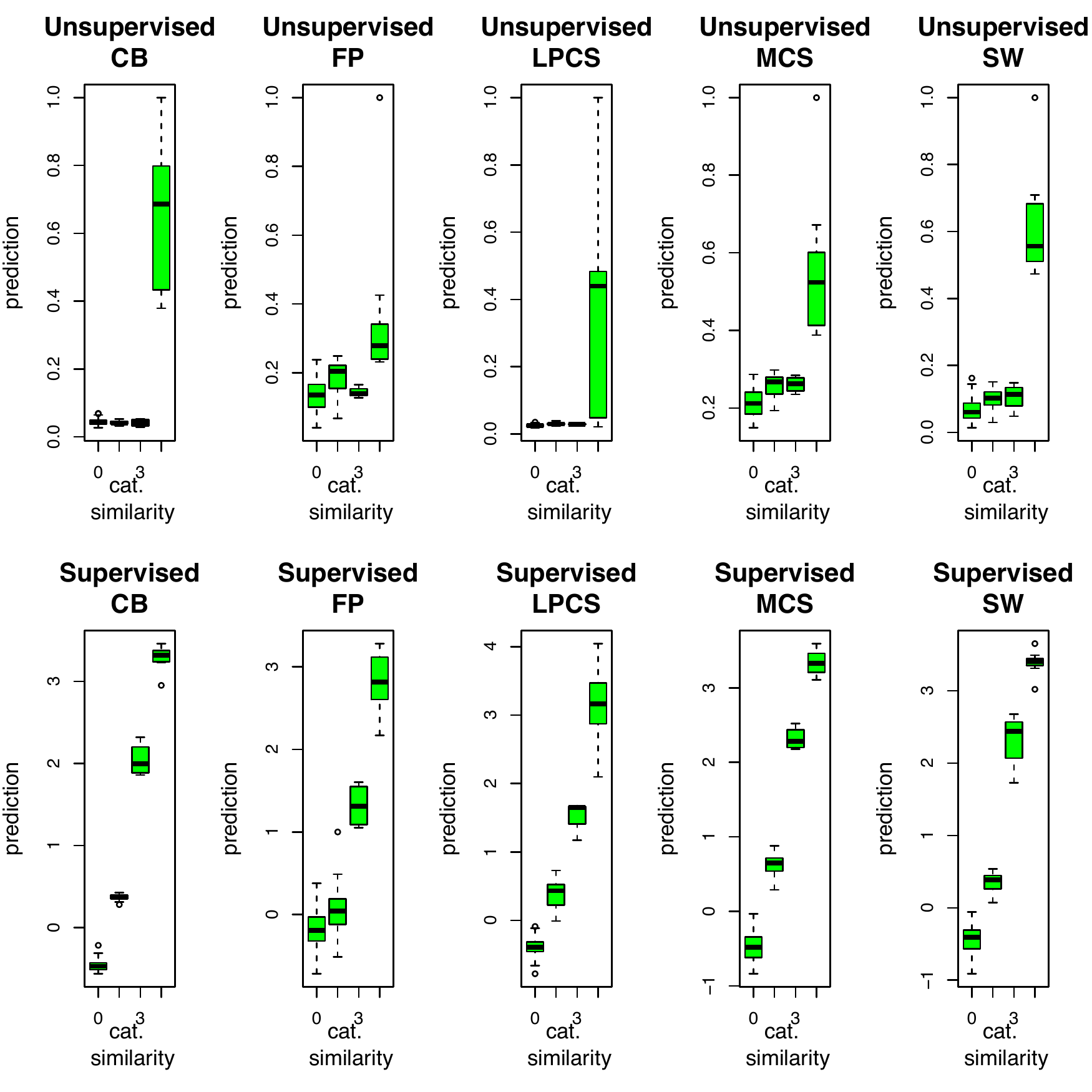}}
\caption{Box-and-whisker plots of the values used for ranking data set II for one randomly chosen query as an example. The different populations on the x-axis denote the groups that are formed by subdividing the database enzymes according to the number of EC number digits they share with the query. Given a query $v$ and a database enzyme $v'$, the y-axis shows the distribution of the values $h(v,v')$ and $K^\phi(v,v')$ for the supervised and the unsupervised approach, respectively. For nearly all cavity-based similarity measures, one can observe a much better separation of the groups for the supervised approach.}\label{box_plots}
\end{figure}

Second, we also advocate that supervised ranking methods have the ability to preserve the hierarchical structure of EC numbers in their predicted rankings. Figure~\ref{test_HM} supports this claim. It summarizes the values used for ranking one fold of the test set obtained by the different models. {A higher value (indicated by a lighter color) in a row means that this enzyme is considered to have a higher catalytic similarity w.r.t.\ the query enzyme.} So, for unsupervised ranking it visualizes $K^\phi(v,v')$, for supervised ranking the values $h(v,v')$ are shown. Each row of the heatmap corresponds to one query. For the supervised models one notices a much better correspondence with the ground truth. Furthermore, the different levels of catalytic similarity can be better distinguished. In addition, an example of the distributions of the predicted values within one query is visualized in Figure~\ref{box_plots} by means of box plots. The different populations within a plot correspond to the different levels of the catalytic similarity w.r.t.\ the query enzyme. {This illustrates again that supervised models can make a better discrimination between enzymes that are functionally more similar and those that are dissimilar}. For this example query, no quartiles are overlapping in any supervised model, unlike the unsupervised approach, which only detects a good ranking for exact matches (i.e., enzymes having an EC number identical to the query). 

A third reason for improvement by the supervised ranking method can be  found in the exploitation of dependencies between different catalytic similarity values. Roughly speaking, if one is interested in the catalytic similarity between enzymes $v$ and $v'$, one can try to compute this catalytic similarity in a direct way based on mutual relationships in cavities, or derive it in an indirect way from the cavity-based similarity with a third enzyme $v''$. This division into a direct and an indirect approach shows a certain correspondence to similar discussions in the context of inferring protein-protein interaction and signal transduction networks -- see e.g.\ ~\cite{Albert2007,Vert2007,Geurts2007}. Unsupervised ranking boils in a certain sense down to a direct approach, while supervised ranking should be interpreted as indirect. Especially when the similarity matrix contains noisy values, one can expect that the indirect approach allows for detecting \emph{back bone} entries and correcting the noisy ones. 


\subsection{Differences between performance measures}

Table~\ref{pal} indicates that the different performances are to some degree influenced by the similarity measure and data set used. This is especially clear for the supervised ranking approach. One can observe a very clear distinction between the ranking accuracy and area under the ROC curve, which treat every position as equally important, and the other two measures, which emphasize the top of the ranking. This should not come as a surprise, as an approximation of the ranking error is optimized in our algorithms. The AUC is very related to the RA, as they coincide for bipartite rankings. In the latter we only make a distinction between 'relevant' enzymes, which have three or more EC numbers in common, and enzymes which do not. Since the RA uses a finer fragmentation of functional similarity, this is a more severe performance measure compared to the AUC. For data set II, both AUC and RA are both very close to the theoretical optimum for nearly all cavity-based similarity measures in the supervised case.

\begin{figure}[t]
   \centering
   \includegraphics[width=0.42\textwidth]{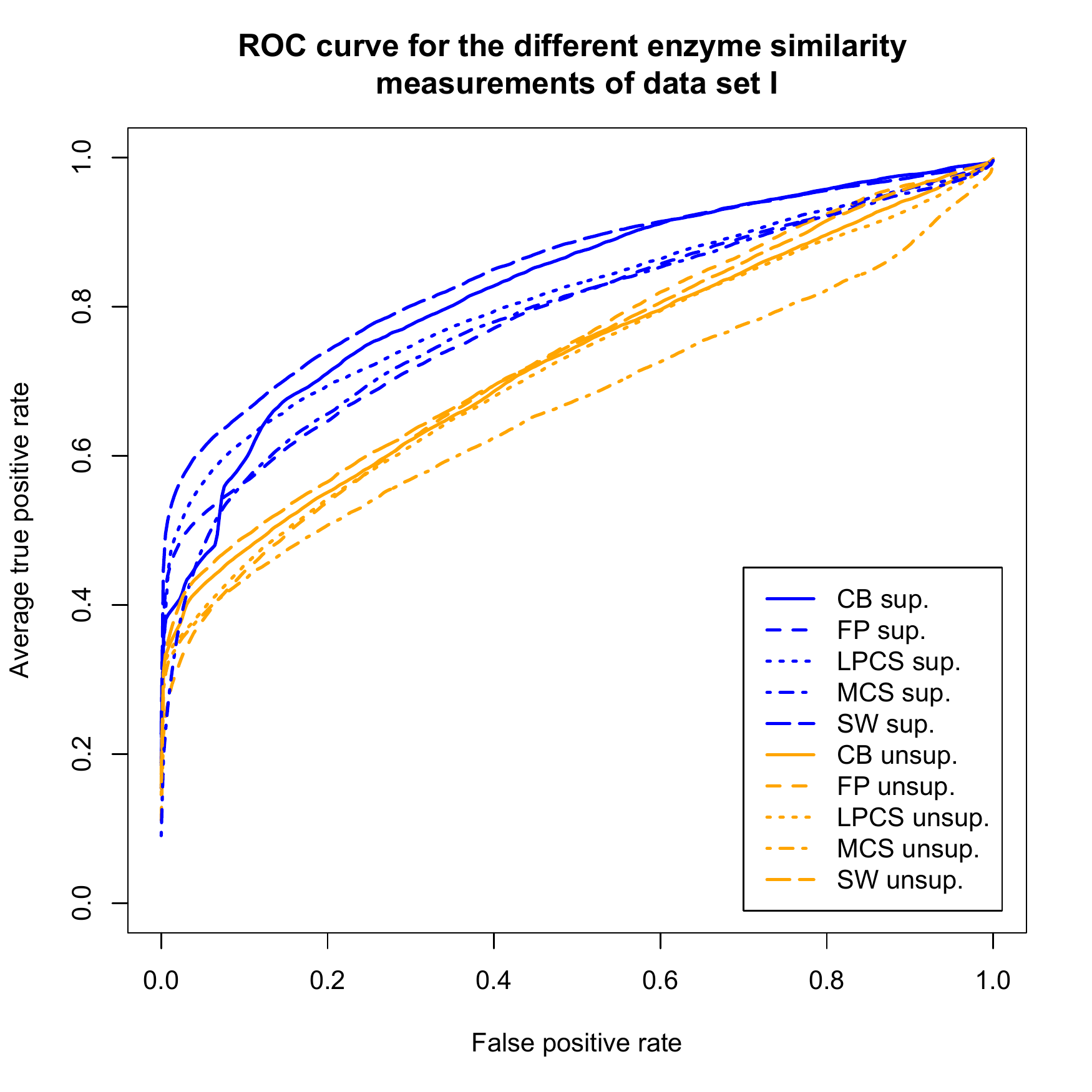} 
   \caption{Receiver operating characteristic curves for the unsupervised and supervised ranking methods for data set I. An enzyme is considered functionally similar to the query if the first three digits of the EC number are identical to those of the query.}
   \label{ROC}
\end{figure}

Figure~\ref{ROC} shows the ROC curves that are obtained by applying the cut-off threshold for data set I, which defines a database enzyme a hit if at least the first three digits of the EC number are correct. In contrast to the scalar performance measures of Table~\ref{pal}, the ROC curve gives information about the quality of the ranking at all positions. It is immediately clear that supervised ranking outcompetes unsupervised ranking, as the former's curves are closer to the upper left corner. Typically for these curves, there is a part in the beginning where the line has a very high slope, showing that a certain fraction of relevant objects can be detected with very high sensitivity and specificity. This fraction that can be detected nearly without mistakes increases after the supervised learning step, as indicated by the higher `offset' of these curves (most clear for SW, LPCS and FP in the ROC curve). The next section of the curve is usually a part with a smaller average slope, indicating that at some point it becomes harder to detect signal. For the unsupervised curves, this is nearly a straight line, which means that from that point the detection of catalytically similar enzymes is essentially random. The supervised curves still have a concave shape at their second part, which shows that relevant enzymes can still be detected in that piece.

From Table~\ref{pal} it is also clear that supervised ranking usually scores worse for MAP and nDCG compared to RA and AUC. For the nDCG the performance sometimes even decreases after learning a model! This can easily be explained by the fact that our model optimizes the quality of the complete ranking, in contrast to only the top which is assessed by MAP and nDGC. Note that the top functionally similar enzymes (i.e. the same EC number) can likely be detected based on the cavity-based similarity alone. Hence, training a model might not be required to perform good in this section. One can see this learning effect nicely in Figure~\ref{ROC} for the CB similarity measure. The quality of the supervised ranking of the top for this measure is worse than for all other measures (indicated by the low nDCG). The overall ranking (indicated by the AUC) is quite good in comparison, as the lower part compensates for the bad ranking at the top. Depending on the application, the top or the general ranking might be more of interest.

\section{Related work}
Since the comparison of enzymes has become an important task in functional bioinformatics, a vast number of similarity measures for proteins has been proposed so far. As mentioned in the introduction, a reliable method will focus on the geometry and the physico-chemical properties of certain regions of an enzyme. However, methods that are based on the sequence or the fold usually exhibit a much lower complexity and they can also lead to good results, especially when the sequence identity is above a certain threshold. \emph{ProFunc}~\cite{Laskowski2005} is in this regard a very interesting tool, in which a bulk of different methods is applied, such as sequence alignment, motif and template search, and also a comparison of active sites. The biological function of enzymes is derived from the closest match in different databases such as PDB, UniProt and PROCAT, and finally returned by the program. Despite being very powerful, this approach becomes nevertheless very inefficient, with runtimes up to several hours fro a single protein. Since we considered a sizeable data set for which nearly 3,000,000 pairwise similarity scores had to be computed, it became impossible to compare our results to ProFunc. 

In addition to focusing on individual enzymes, one can also take protein-protein interactions into account for inferring the function. Proteins that are close to each other in a protein-protein interaction network are expected to have similar functions, so one can try to infer the function of an unanotated protein by looking at its neighbors~\cite{Schwikowski2000}. Similarly, one can also solve optimization problems over the global network, such as maximizing the number of edges that connect proteins sharing the same function~\cite{Vazquez2003, Karaoz2004}. Other approaches make use of probabilistic graphical models such as Markov random fields~\cite{Deng2003, Letovsky2003}.
Conceptually, these methods might also enrich the predictions obtained by an unsupervised search-based approach, but they usually do not consider cavity and binding site information to predict the function of proteins. 

\section{Conclusion}

In this paper we have recast the EC annotation problem as a conditional ranking problem. We have shown that retrieval of enzymes w.r.t.\ functionality can be substantially improved by applying a supervised ranking method that takes advantage of ground truth EC numbers during the training phase. In contrast, more traditional methods rely heavily on a notion of similarity to search for functionally related enzymes in online databases. Such methods lead to an unsupervised approach in which annotations are not taken into account.

We focused specifically on cavity-based similarity measures, because their benefits compared to sequence-based and structure-based approaches have been demonstrated in previous work, although our method can work with any meaningful similarity measure defined on enzymes. In the experiments we could demonstrate a considerable
 improvement of the quality of the overall ranking. The results were influenced by the type of data used and the way the ranking was evaluated, indicating that the most optimal method is highly dependent of the specific problem setting. Nevertheless, our supervised ranking algorithm outperformed the unsupervised ranking algorithm for all cavity-based similarities for most performance measures considered. While the unsupervised approach succeeded quite well in returning exact matches to a query, the hierarchical structure of EC numbers was better preserved in the rankings predicted by the supervised approach. As such, supervised ranking can be interpreted as a powerful {alternative} for retrieval methods that are more traditionally used in bioinformatics.

\section*{Acknowledgements}
MS, BDB and WW acknowledge the support of Ghent University (MRP Bioinformatics: from nucleotides to networks). T.F., E.H., S.G.\ and G.K. gratefully acknowledge financial support by the German Research Foundation (DFG) and the LOEWE Research Center for Synthetic Microbiology, Marburg. T.P.\ and A.A.\ are both supported for this work by the Academy of Finland (grant 134020 and 128061, respectively).

\bibliography{bib}

\begin{thebibliography}{10}

\bibitem{Friedberg2006}
I.~Friedberg, ``{Automated protein function prediction--the genomic
  challenge},'' {\em Briefings in Bioinformatics}, vol.~7, pp.~225--42, Sept.
  2006.

\bibitem{Dunaway-Mariano2008}
D.~Dunaway-Mariano, ``{Enzyme function discovery},'' {\em Structure}, vol.~16,
  no.~11, pp.~1599--1600, 2008.

\bibitem{Altschul1990}
S.~F. Altschul, W.~Gish, W.~Miller, E.~W. Myers, and D.~J. Lipman, ``{Basic
  Local Alignment Search Tool},'' {\em Journal of Molecular Biology}, vol.~215,
  no.~3, pp.~403--410, 1990.

\bibitem{Altschul1997}
S.~F. Altschul, T.~L. Madden, A.~A. Sch\"{a}ffer, J.~Zhang, Z.~Zhang,
  W.~Miller, and D.~J. Lipman, ``{Gapped Blast and PsiBlast: a new generation
  of protein database search programs},'' {\em Nucleic Acid Research}, vol.~25,
  no.~17, pp.~3389--3402, 1997.

\bibitem{Leslie2002}
C.~Leslie, E.~Eskin, and W.~S. Noble, ``{The spectrum kernel: a string kernel
  for SVM protein classification.},'' in {\em Pacific Symposium on
  Biocomputing}, pp.~566--575, Jan. 2002.

\bibitem{Leslie2004}
C.~S. Leslie, E.~Eskin, A.~Cohen, J.~Weston, and W.~S. Noble, ``{Mismatch
  string kernels for discriminative protein classification},'' {\em
  Bioinformatics}, vol.~20, pp.~467--76, Mar. 2004.

\bibitem{Powers2006}
R.~Powers, J.~C. Copeland, K.~Germer, K.~A. Mercier, V.~Ramanathan, and
  P.~Revesz, ``Comparison of protein active site structures for functional
  annotation of proteins and drug design,'' {\em Proteins: Structure, Function,
  and Bioinformatics}, vol.~65, pp.~124--135, 2006.

\bibitem{Rost2002}
B.~Rost, ``{Enzyme function less conserved than anticipated},'' {\em Journal of
  Molecular Biology}, vol.~318, no.~2, pp.~595--608, 2002.

\bibitem{Chalk2004}
A.~J. Chalk, C.~L. Worth, J.~P. Overington, and A.~W.~E. Chan, ``{PDBLIG:
  C}lassification of small molecular protein binding in the protein data
  bank,'' {\em Journal of Medical Chemistry}, vol.~47, no.~15, pp.~3807--3816,
  2004.

\bibitem{Kinoshita2007}
K.~Kinoshita, Y.~Murakami, and H.~Nakamura, ``{eF}-seek: prediction of the
  functional sites of proteins by searching for similar electrostatic potential
  and molecular surface shape,'' {\em Nucleic Acid Research}, vol.~35,
  no.~suppl 2, pp.~W398--W402, 2007.

\bibitem{Thornton2000}
J.~Thornton, A.~Todd, D.~Milburn, N.~Borkakoti, and C.~Orengo, ``{From
  structure to function: Approaches and limitations},'' {\em Nature Structural
  Biology}, vol.~7 Suppl, pp.~991--994, 2000.

\bibitem{Kotera2004a}
M.~Kotera, Y.~Okuno, M.~Hattori, S.~Goto, and M.~Kanehisa, ``{Computational
  assignment of the EC numbers for genomic-scale analysis of enzymatic
  reactions},'' {\em Journal of the American Chemical Society}, vol.~126,
  pp.~16487--98, Dec. 2004.

\bibitem{Egelhofer2010}
V.~Egelhofer, I.~Schomburg, and D.~Schomburg, ``{Automatic assignment of EC
  numbers},'' {\em PLoS Computational Biology}, vol.~6, pp.~1000661+, 2010.

\bibitem{Shatsky2004}
M.~Shatsky, R.~Niussinov, and H.~J. Wolfson, ``A method for simultaneous
  alignment of multiple protein structures,'' {\em Proteins: Structure,
  Function, and Bioinformatics}, vol.~56, pp.~143--156, 2004.

\bibitem{Shatsky2006a}
M.~Shatsky, {\em The Common Point Set Problem with Applications to Protein
  Structure Analysis}.
\newblock PhD thesis, Tel Aviv University, Tel Aviv, Israel, 2006.

\bibitem{Harrison2003}
A.~Harrison, F.~Pearl, I.~Sillitoe, T.~Slidel, R.~Mott, J.~Thornton, and
  C.~Orengo, ``{Recognizing the fold of a protein structure},'' {\em
  Bioinformatics}, vol.~19, no.~14, pp.~1748--1759, 2003.

\bibitem{Martin1998}
A.~C. Martin, C.~A. Orengo, E.~G. Hutchinson, S.~Jones, M.~Karmirantzou, R.~A.
  Laskowski, J.~B. Mitchell, C.~Taroni, and J.~M. Thornton, ``{Protein folds
  and functions},'' {\em Structure}, vol.~6, pp.~875--84, July 1998.

\bibitem{Babbitt1997}
P.~C. Babbitt and J.~Gerlt, ``{Understanding enzyme superfamilies},'' {\em
  Journal of Biological Chemistry}, vol.~272, pp.~30591--30594, Dec. 1997.

\bibitem{Gerlt2001}
J.~A. Gerlt and P.~C. Babbitt, ``{Divergent evolution of enzymatic function:
  mechanistically diverse superfamilies and functionally},'' {\em Annual review
  of biochemistry}, vol.~70, pp.~209--246, 2001.

\bibitem{Osadchy2011}
M.~Osadchy and R.~Kolodny, ``{Maps of protein structure space reveal a
  fundamental relationship between protein structure and function},'' {\em
  Proceedings of the National Academy of Sciences of the United States of
  America}, vol.~108, no.~30, pp.~12301--12306, 2011.

\bibitem{Kristensen2008}
D.~M. Kristensen, R.~M. Ward, A.~M. Lisewski, S.~Erdin, B.~Y. Chen, V.~Y.
  Fofanov, M.~Kimmel, L.~E. Kavraki, and O.~Lichtarge, ``{Prediction of enzyme
  function based on 3D templates of evolutionarily important amino acids},''
  {\em BMC Bioinformatics}, vol.~9, Jan. 2008.

\bibitem{Erdin2011}
S.~Erdin, A.~M. Lisewski, and O.~Lichtarge, ``{Protein function prediction:
  towards integration of similarity metrics},'' {\em Current Opinion in
  Structural Biology}, vol.~21, pp.~180--8, Apr. 2011.

\bibitem{1996Laskowski}
R.~A. Laskowski, N.~M. Luscombe, M.~B. Swindells, and J.~M. Thornton,
  ``{Protein clefts in molecular recognition and function},'' {\em Protein
  Science}, vol.~5, no.~12, pp.~2438--2452, 1996.

\bibitem{Perot10}
S.~P\'{e}rot, O.~Sperandio, M.~A. Miteva, A.-C. Camproux, and B.~O.
  Villoutreix, ``{Druggable pockets and binding site centric chemical space: a
  paradigm shift in drug discovery},'' {\em Drug Discovery Today}, vol.~15,
  no.~15-16, pp.~656--667, 2010.

\bibitem{Andersson2010}
C.~Andersson, B.~Y. Chen, and A.~Linusson, ``{Mapping of ligand-binding
  cavities in proteins},'' {\em Proteins}, vol.~78, no.~6, pp.~1408--1422,
  2010.

\bibitem{Weisel2010}
M.~Weisel, J.~M. Kriegl, and G.~Schneider, ``{Architectural repertoire of
  ligand-binding pockets on protein surfaces},'' {\em Chembiochem}, vol.~11,
  no.~4, pp.~556--563, 2010.

\bibitem{Weber2004}
A.~Weber, A.~Casini, A.~Heine, D.~Kuhn, C.~T. Supuran, A.~Scozzafava, and
  G.~Klebe, ``{Unexpected nanomolar inhibition of carbonic anhydrase by
  COX-2-selective celecoxib: new pharmacological opportunities due to related
  binding site recognition},'' {\em Journal of Medical Chemistry}, vol.~47,
  no.~3, pp.~550--557, 2004.

\bibitem{Huan1999}
J.~Huan, D.~Bandyopadhyay, W.~Wang, J.~Snoeyink, J.~Prins, and A.~Tropsha,
  ``{Comparing graph representations of protein structure for mining
  family-specific residue-based packing motifs},'' {\em Journal of
  computational biology : a journal of computational molecular cell biology},
  vol.~12, no.~6, pp.~657--71, 1999.

\bibitem{Borgwardt2005}
K.~M. Borgwardt, C.~S. Ong, S.~Sch\"{o}nauer, S.~V.~N. Vishwanathan, A.~J.
  Smola, and H.-P. Kriegel, ``{Protein function prediction via graph
  kernels},'' {\em Bioinformatics}, vol.~21 Suppl 1, pp.~i47--56, June 2005.

\bibitem{Gartner2008}
T.~G\"artner, {\em Kernels for Structured Data}.
\newblock Singapore: World Scientific, 2008.

\bibitem{Shervashidze2009}
N.~Shervashidze, ``{Efficient graphlet kernels for large graph comparison},''
  in {\em Proceedings of the 12th International Conference on Artificial
  Intelligence and Statistics (AISTATS)}, vol.~5, pp.~488--495, 2009.

\bibitem{Vacic2010}
V.~Vacic, L.~I. ans S.~Lonardi, and P.~Radivojac, ``Graphlet kernels for
  prediction of functional residues in protein structures,'' {\em Journal of
  Computational Biology}, vol.~17, no.~1, pp.~55--72, 2010.

\bibitem{Shatsky2006}
M.~Shatsky, A.~Shulman-Peleg, R.~Nussinov, and H.~J. Wolfson, ``The multiple
  common point set problem and its application to molecule binding pattern
  detection,'' {\em Journal of Computational Biology}, vol.~13, no.~2,
  pp.~407--428, 2006.

\bibitem{Fober2011}
T.~Fober, S.~Glinca, G.~Klebe, and E.~H\"ullermeier, ``Superposition and
  alignment of labeled point clouds,'' {\em IEEE/ACM Transactions on
  Computational Biology and Bioinformatics}, vol.~8, pp.~1653--1666,
  November/December 2011.

\bibitem{Weill2010}
N.~Weill and D.~Rognan, ``Alignment-free ultra-high-throughput comparison of
  druggable protein-ligand binding sites,'' {\em Journal of Chemical
  Information and Modeling}, vol.~50, no.~1, pp.~123--135, 2010.

\bibitem{Fober2012}
T.~Fober, M.~Mernberger, G.~Klebe, and E.~H\"ullermeier, ``Fingerprint kernels
  for protein structure comparison,'' {\em Molecular Informatics}, vol.~31,
  no.~6-7, pp.~443--452, 2012.

\bibitem{Hullermeier2010a}
E.~H{\"ullermeier} and J.~F{\"u}rnkranz, {\em Preference Learning}.
\newblock Springer, 2010.

\bibitem{Rathke:2010cz}
F.~Rathke, K.~Hansen, U.~Brefeld, and K.-R. M\"{u}ller, ``{StructRank: a new
  approach for ligand-based virtual screening},'' {\em Journal of Chemical
  Information and Modeling}, vol.~51, pp.~83--92, Dec. 2010.

\bibitem{Agarwal2010a}
S.~Agarwal, D.~Dugar, and S.~Sengupta, ``{Ranking chemical structures for drug
  discovery: a new machine learning approach.},'' {\em Journal of Chemical
  Information and Modeling}, vol.~50, pp.~716--31, May 2010.

\bibitem{Weston2004}
J.~Weston, A.~Eliseeff, D.~Zhou, C.~Leslie, and W.~S. Noble, ``{Protein
  ranking: from local to global structure in the protein similarity network},''
  {\em Proceedings of the National Academy of Science}, vol.~101,
  pp.~6559--6563, 2004.

\bibitem{Kuang2005}
R.~Kuang, J.~Weston, W.~S. Noble, and C.~Leslie, ``{Motif-based protein ranking
  by network propagation},'' {\em Bioinformatics}, vol.~21, pp.~3711--8, Oct.
  2005.

\bibitem{Pahikkala2009}
T.~Pahikkala, E.~Tsivtsivadze, A.~Airola, J.~J\"{a}rvinen, and J.~Boberg, ``{An
  efficient algorithm for learning to rank from preference graphs},'' {\em
  Machine Learning}, vol.~75, no.~1, pp.~129--165, 2009.

\bibitem{Joachims2005}
T.~Joachims, ``A support vector method for multivariate performance measures,''
  in {\em Proceedings of the International Conference on Machine Learning,
  Bonn, Germany}, pp.~377--384, 2005.

\bibitem{Dobson2005a}
P.~D. Dobson and A.~J. Doig, ``{Predicting enzyme class from protein structure
  without alignments.},'' {\em Journal of Molecular Biology}, vol.~345,
  pp.~187--99, Jan. 2005.

\bibitem{Rousu2006}
J.~Rousu, C.~Saunders, S.~Szedmak, and J.~Shawe-Taylor, ``{Kernel-based
  learning of hierarchical multilabel classification models},'' {\em Journal of
  Machine Learning Research}, vol.~7, pp.~1601--1626, 2006.

\bibitem{Sokolov2008}
A.~Sokolov and A.~Ben-Hur, ``{A structured-outputs method for prediction of
  protein function},'' in {\em Proceedings of the 3rd International Workshop on
  Machine Learning in Systems Biology}, 2008.

\bibitem{Arakaki2009}
A.~K. Arakaki, Y.~Huang, and J.~Skolnick, ``{EFICAz2: enzyme function inference
  by a combined approach enhanced by machine learning},'' {\em BMC
  Bioinformatics}, vol.~10, p.~107, Jan. 2009.

\bibitem{Hendlich2003}
M.~Hendlich, A.~Bergner, J.~G\"{u}nther, and G.~Klebe, ``{Relibase: design and
  development of a database for comprehensive analysis of protein–ligand
  interactions},'' {\em Journal of Molecular Biology}, vol.~326, pp.~607--620,
  Feb. 2003.

\bibitem{Schmitt2002}
S.~Schmitt, D.~Kuhn, and G.~Klebe, ``{A new method to detect related function
  among proteins independent of sequence and fold homology},'' {\em Journal of
  Molecular Biology}, vol.~323, no.~2, pp.~387--406, 2002.

\bibitem{Bunke1998}
H.~Bunke and K.~Shearer, ``{A graph distance metric based on the maximal common
  subgraph},'' {\em Pattern Recognition Letters}, vol.~19, no.~3-4,
  pp.~255--259, 1998.

\bibitem{Sanfeliu1983}
A.~Sanfeliu and K.~Fu, ``{A distance measure between attributed relational
  graphs for pattern recognition},'' {\em {IEEE} Transactions on Systems, Man
  and Cybernetics}, vol.~13, no.~3, pp.~353--362, 1983.

\bibitem{Weskamp2007}
N.~Weskamp, E.~H\"{u}llermeier, D.~Kuhn, and G.~Klebe, ``{Multiple graph
  alignment for the structural analysis of protein active sites.},'' {\em
  IEEE/ACM Transactions on Computational Biology and Bioinformatics}, vol.~4,
  no.~2, pp.~310--20, 2007.

\bibitem{Neuhaus2007}
M.~Neuhaus and H.~Bunke, {\em Bridging the Gap between Graph Edit Distance and
  Kernel Machines}.
\newblock New Jersey: World Scientific, 2007.

\bibitem{Borgwardt2005a}
K.~Borgwardt, C.~Ong, S.~Schonauer, S.~Vishwanathan, A.~Smola, and H.~Kriegel,
  ``{Protein function prediction via graph kernels},'' {\em Bioinformatics},
  vol.~21, no.~1, pp.~i47--i56, 2005.

\bibitem{Alt1996}
H.~Alt and L.~J. Guibas, ``Discrete geometric shapes: Matching, interpolation,
  and approximation: A survey,'' tech. rep., Handbook of Computational
  Geometry, 1996.

\bibitem{Mahe2005}
P.~Mah\'{e}, N.~Ueda, T.~Akutsu, J.-L. Perret, and J.-P. Vert, ``{Graph kernels
  for molecular structure-activity relationship analysis with support vector
  machines},'' {\em Journal of chemical information and modeling}, vol.~45,
  no.~4, pp.~939--51, 2005.

\bibitem{Deza2009}
M.~M. Deza and E.~Deza, {\em Encyclopedia of Distances}.
\newblock Heidelberg, Germany: Springer, 2009.

\bibitem{Beyer2002}
H.-G. Beyer and H.-P. Schwefel, ``Evolution strategies: {A} comprehensive
  introduction,'' {\em Natural Computing}, vol.~1, no.~1, pp.~3--52, 2002.

\bibitem{Hoffmann2010}
B.~Hoffmann, M.~Zaslavskiy, J.-P. Vert, and V.~Stoven, ``{A new protein binding
  pocket similarity measure based on comparison of clouds of atoms in 3D:
  application to ligand prediction},'' {\em BMC Bioinformatics}, vol.~11, Jan.
  2010.

\bibitem{Fober2009}
T.~Fober, M.~Mernberger, G.~Klebe, and E.~H\"{u}llermeier, ``{Evolutionary
  construction of multiple graph alignments for the structural analysis of
  biomolecules.},'' {\em Bioinformatics}, vol.~25, pp.~2110--7, Aug. 2009.

\bibitem{Kabsch1976}
W.~Kabsch, ``A solution of the best rotation to relate two sets of vectors,''
  {\em Acta Crystallographica}, vol.~32, pp.~922--923, 1976.

\bibitem{Fober2009b}
T.~Fober, M.~Mernberger, R.~Moritz, and E.~H\"ullermeier, ``Graph-kernels for
  the comparative analysis of protein active sites,'' in {\em German Conference
  on Bioinformatics}, (Halle (Saale), Germany), pp.~21 -- 31, 2009.

\bibitem{Smith1981}
T.~F. Smith and M.~S. Waterman, ``{Identification of common molecular
  subsequences},'' {\em Journal of Molecular Biology}, vol.~147, pp.~195--197,
  1981.

\bibitem{Pahikkala2010}
T.~Pahikkala, W.~Waegeman, E.~Tsivtsivadze, T.~Salakoski, and B.~{De Baets},
  ``{Learning intransitive reciprocal relations with kernel methods},'' {\em
  European Journal of Operational Research}, vol.~206, pp.~676--685, Nov. 2010.

\bibitem{pahikkala2013conditional}
T.~Pahikkala, A.~Airola, M.~Stock, B.~{De Baets}, and W.~Waegeman, ``{Efficient
  regularized least-squares algorithms for conditional ranking on relational
  data},'' {\em Machine Learning}, vol.~93, no.~2-3, pp.~321--356, 2013.

\bibitem{Scholkopf2002}
B.~Sch\"olkopf and A.~Smola, {\em Learning with Kernels, Support Vector
  Machines, Regularisation, Optimization and Beyond}.
\newblock The MIT Press, 2002.

\bibitem{Waegeman2012}
W.~Waegeman, T.~Pahikkala, A.~Airola, T.~Salakoski, M.~Stock, and B.~{De
  Baets}, ``{A kernel-based framework for learning graded relations from
  data},'' {\em IEEE Transactions on Fuzzy Systems}, vol.~20, pp.~1090--1101,
  2012.

\bibitem{Ben-Hur2005}
A.~Ben-Hur and W.~S. Noble, ``{Kernel methods for predicting protein-protein
  interactions},'' {\em Bioinformatics}, vol.~21, pp.~i38--46, June 2005.

\bibitem{kashima2009}
H.~Kashima, S.~Oyama, Y.~Yamanishi, and K.~Tsuda, ``On pairwise kernels: An
  efficient alternative and generalization analysis.,'' in {\em PAKDD}
  (T.~Theeramunkong, B.~Kijsirikul, N.~Cercone, and T.~B. Ho, eds.), vol.~5476
  of {\em Lecture Notes in Computer Science}, pp.~1030--1037, Springer, 2009.

\bibitem{Vert2007}
J.-P. Vert, J.~Qiu, and W.~S. Noble, ``{A new pairwise kernel for biological
  network inference with support vector machines.},'' {\em BMC Bioinformatics},
  vol.~8, Jan. 2007.

\bibitem{Herbrich2000}
R.~Herbrich, T.~Graepel, and K.~Obermayer, ``Large margin rank boundaries for
  ordinal regression,'' in {\em Advances in Large Margin Classifiers}
  (A.~Smola, P.~Bartlett, B.~Sch\"olkopf, and D.~Schuurmans, eds.),
  pp.~115--132, MIT Press, 2000.

\bibitem{Chapelle2006}
O.~Chapelle, B.~Sch\"{o}lkopf, and A.~Zien, {\em {Semi-Supervised Learning}}.
\newblock MIT Press, 2006.

\bibitem{Waegeman2008a}
W.~Waegeman, B.~{De Baets}, and L.~Boullart, ``{ROC analysis in ordinal
  regression learning},'' {\em Pattern Recognition Letters}, vol.~29, pp.~1--9,
  2008.

\bibitem{Jarvelin2002}
K.~J\"{a}rvelin and J.~Kek\"{a}l\"{a}inen, ``{Cumulated gain-based evaluation
  of IR techniques},'' {\em ACM Transactions on Information Systems (TOIS)},
  vol.~20, no.~4, pp.~422--446, 2002.

\bibitem{Varma2006}
S.~Varma and R.~Simon, ``{Bias in error estimation when using cross-validation
  for model selection},'' {\em BMC Bioinformatics}, vol.~7, Jan. 2006.

\bibitem{Albert2007}
R.~Albert, B.~DasGupta, R.~Dondi, S.~Kachalo, E.~Sontag, A.~Zelikovsky, and
  K.~Westbrooks vol.~14, pp.~927--949, 2007.

\bibitem{Geurts2007}
P.~Geurts, N.~Touleimat, M.~Dutreix, and F.~D'Alch\'{e}-Buc, ``{Inferring
  biological networks with output kernel trees},'' {\em BMC Bioinformatics},
  vol.~8, no.~2, p.~S4, 2007.

\bibitem{Laskowski2005}
R.~Laskowski, J.~Watson, and J.~Thornton, ``Profunc: a server for predicting
  protein function from 3d stucture,'' {\em NAR}, vol.~33, pp.~W89--W93, 2005.

\bibitem{Schwikowski2000}
B.~Schwikowski, P.~Uetz, and S.~Fields, ``A network of protein-protein
  interactions in yeast,'' {\em Nature Biotechnology}, vol.~18, pp.~3548--3557,
  2000.

\bibitem{Vazquez2003}
A.~Vazquez, A.~Flammini, A.~Maritan, and A.~Vespignani, ``Global protein
  function prediction from protein-protein interaction networks,'' {\em
  Bioinformatics}, vol.~21, pp.~ii59--ii65, 2003.

\bibitem{Karaoz2004}
U.~Karaoz, T.~Murali, S.~Letovsky, Y.~Zheng, C.~Ding, C.~Cantor, and S.~Kasif,
  ``Whole-genome annotation by using evidence integration in functional-linkage
  networks,'' {\em Proceedings of the National Academy of Sciences}, vol.~101,
  pp.~2888--2893, 2004.

\bibitem{Deng2003}
M.~Deng, K.~Zhang, S.~Mehta, T.~Chen, and F.~Sun, ``Prediction of protein
  function using protein-protein interaction data,'' {\em Journal of
  Computational Biology}, vol.~10, pp.~947--960, 2003.

\bibitem{Letovsky2003}
S.~Letovsky and S.~Kasif, ``Predicting protein function from protein/protein
  interaction data: a probabilistic approach,'' {\em Bioinformatics}, vol.~19,
  pp.~i197--i204, 2003.

\end{thebibliography}

\end{document}